\documentclass[conference]{IEEEtran}

\usepackage{cite}
\usepackage{amsmath,amssymb,amsfonts}
\usepackage{algorithmic}
\usepackage{graphicx}
\usepackage{textcomp}
\usepackage{xcolor}
\def\BibTeX{{\rm B\kern-.05em{\sc i\kern-.025em b}\kern-.08em
    T\kern-.1667em\lower.7ex\hbox{E}\kern-.125emX}}
\begin{document}

\title{Deep Learning for Spatiotemporal Big Data: A Vision on Opportunities and Challenges
}

\author{
\IEEEauthorblockN{Zhe Jiang}
\IEEEauthorblockA{\textit{Department of Computer \& Information Science \& Engineering} \\
\textit{University of Florida, Gainesville, FL, USA
}\\
zhe.jiang@ufl.edu}
}

\maketitle

\begin{abstract}
With advancements in GPS, remote sensing, and computational simulation, an enormous volume of spatiotemporal data is being collected at an increasing speed from various application domains, spanning Earth sciences, agriculture, smart cities, and public safety. Such emerging geospatial and spatiotemporal big data, coupled with recent advances in deep learning technologies, foster new opportunities to solve problems that have not been possible before. For instance, remote sensing researchers can potentially train a foundation model using Earth imagery big data for numerous land cover and land use modeling tasks. Coastal modelers can train AI surrogates to speed up numerical simulations. However, the distinctive characteristics of spatiotemporal big data pose new challenges for deep learning technologies. This vision paper introduces various types of spatiotemporal big data, discusses new research opportunities in the realm of deep learning applied to spatiotemporal big data, lists the unique challenges, and identifies several future research needs.

\end{abstract}


\section{Introduction}
With advancements in GPS, remote sensing, and computational simulation, a rapidly expanding volume of spatiotemporal data is being collected from diverse application domains, ranging from Earth sciences to agriculture, smart cities to public safety~\cite{eftelioglu2017nexus}. For instance, NASA's Earth Observing System Data and Information System (EOSDIS) alone houses 90 petabytes of Earth science data that continues to grow daily~\cite{nasadata}. These emerging spatiotemporal data, characterized by their volume, velocity, and variety, which surpass the capabilities of traditional computational platforms, are referred to as \emph{spatiotemporal big data} (STBD)~\cite{jiang2017spatialbig}.

Analyzing geospatial and spatiotemporal data has attracted researchers' interest since early 2000, fostering the development of spatial and spatiotemporal data mining~\cite{shekhar2015spatiotemporal,atluri2018spatio,jiang2019survey,sharma202221}, a research field that aims to discover non-trivial, previously unknown, but potentially useful patterns or make accurate predictions on massive geospatial and spatiotemporal data. In this context, the word ``massive" refers to a data size that exceeds the capability of traditional statistical methodologies (geostatistics, time series analysis)~\cite{cressie2015statistics}, necessitating efficient algorithms such as computational pruning. The rapid growth of this interdisciplinary field not only enhances the awareness of available tools for domain scientists to make sense of their spatiotemporal data but also inspires data mining researchers to develop novel algorithms that address unique characteristics of spatiotemporal data, where samples violate the common independent and identical distribution (i.i.d.) assumption. 

The emergence of STBD in the last decade has inspired the development of spatial big data computational platforms (e.g., Hadoop-GIS~\cite{aji2013hadoop}, SpatialHadoop~\cite{eldawy2015spatialhadoop}, GeoSpark~\cite{yu2015geospark}). These platforms make it possible to analyze patterns in STBD that are beyond the capability of traditional spatiotemporal data mining algorithms. However, most of these systems focus on computational scalability in simple query processing or visual pattern analysis, instead of complex predictive learning tasks.

The recent advances in deep learning technologies have brought unique opportunities. Thanks to the availability of big training data (e.g., ImageNet), efficient computational architectures (e.g., GPUs), and smarter models and learning algorithms, deep learning has achieved tremendous success in computer vision and language processing applications~\cite{lecun2015deep}. It has already made promising progress in spatial and spatiotemporal domains such as Earth system process modeling~\cite{reichstein2019deep}, drug discovery~\cite{jumper2021highly}, transportation and urban planning~\cite{yin2021deep}. 

When STBD meets deep learning, new opportunities arise. First, deep neural networks can provide end-to-end training without handcrafting spatial and temporal contextual features~\cite{reichstein2019deep}. Traditionally, in spatiotemporal data mining, a major step is feature engineering based on domain knowledge. This approach has limitations. For instance, researchers have to repeat this tedious and time-consuming process for every new application, and the handcrafted features may be insufficient to capture complex spatial and temporal patterns. Second, some complex spatiotemporal patterns are only visible in STBD and thus are not detectable by traditional spatial and spatiotemporal data mining methods. In contrast, deep learning models are very data-hungry with a large number of model parameters (e.g., millions to billions). The huge capacity of deep learning models provides opportunities to learn hidden patterns from STBD that are otherwise ignored by traditional models. 

This vision paper will briefly discuss the emerging opportunities of deep learning for spatiotemporal big data, highlight the unique challenges, and identify new research directions for this highly interdisciplinary area.

\section{Types of STBD}
\subsection{Spatial data}

\emph{Point reference data}: Spatial point reference data record non-spatial attribute values at specific point locations in continuous space. The assumption is that the attribute exists everywhere in the spatial domain but is only observed at these sampled point locations. For example, this could include the precipitation measurements obtained from various weather stations, each representing a point location.

\emph{Areal data}: Areal data capture non-spatial attribute values across the entire continuous space by dividing it into cells, either regular or irregular, and assigning an aggregated value to each cell. An example of areal data is a map displaying the number of COVID-19 cases in all U.S. counties.

\emph{Spatial point process}: Spatial point processes involve the recording of spatial point locations where specific types of events occur, such as crimes or traffic accidents. Unlike point reference data, the locations of events in a spatial point process are treated as random variables rather than non-spatial attributes. Examples include the spatial distribution of traffic crash locations or crime events.

\emph{Spatial network}: Spatial networks are graphs in which nodes and edges correspond to spatial objects. A common example is a road network, where nodes represent road intersections, and edges represent road segments.

\subsection{Spatiotemporal data}
When considering the additional time dimension on top of spatial data, we can divide the time dimension into two cases: regular time intervals (discrete time) or irregular time intervals (continuous time). Based on this categorization, we can further identify the following common types of spatiotemporal data.

\emph{Spatiotemporal point reference data}: Spatiotemporal point reference data expand upon point reference data by including non-spatiotemporal attributes sampled in both continuous space and continuous time. An example is water quality samples collected by boats at various locations and timestamps.

\emph{Spatial point time series}: Spatial point time series extend point reference data by incorporating regular time intervals for sampling non-spatial attributes. For instance, hourly temperature measurements from a network of weather stations, with spatial locations sampled in continuous space, fall into this category.

\emph{Areal temporal snapshots (or time series)}: Areal temporal snapshots or time series add a time dimension to areal models, with observations at irregular or regular time intervals. For example, ecologists might collect Earth imagery during different seasons (e.g., leaf-on and leaf-off) to monitor changes in wetland locations.

\emph{Spatiotemporal point process}: Spatiotemporal point processes extend spatial point processes by characterizing both the location and time of point events. An example is traffic crash records, which include information about both the location and time of each incident.

\emph{Trajectories}: Trajectories represent a specialized case of spatiotemporal point processes where points are regularly sampled in time, and the locations follow a sequential order. This is often seen in tracking systems for vehicles or individuals.

\emph{Spatial network time series}: Spatial network time series expand upon spatial network data by including time series data for non-spatiotemporal attributes associated with network nodes or edges. For instance, sensors monitoring traffic speed and flow volume at different road segments provide insights into transportation dynamics over time.

\section{New Opportunities of Deep Learning for Spatiotemporal Big Data}

\subsection{Why is deep learning for STBD different?}



There are several aspects that make deep learning transformative compared with traditional spatial and spatiotemporal data mining techniques for STBD. 

First, deep neural networks enable end-to-end training, eliminating the need for manually crafting spatial contextual features. In traditional spatial and spatiotemporal machine learning, a significant step involves handcrafting these features based on domain knowledge. Deep learning mitigates this by learning relevant features from data directly. This approach offers advantages such as reducing the tedium and time consumption associated with feature engineering for each new application. For instance, in Earth science applications like land cover mapping, deep convolutional neural networks can learn rich spatial and spectral signatures directly from remote sensing image pixels.

Second, deep learning models, characterized by their extensive number of parameters (often ranging from millions to billions), are data-hungry and possess a significantly larger capacity to uncover intricate spatial and temporal patterns. This is particularly crucial for understanding complex spatiotemporal processes involving multiple interacting physical variables operating at various spatial and temporal scales. Deep neural networks are well-suited to capture such complex patterns from STBD.

Third, the recent success of large language models~\cite{brown2020language}, also known as foundation models, in language processing tasks offers insights into addressing a persistent challenge in spatiotemporal data mining: model generalizability across heterogeneous space and time. Traditionally, due to spatial heterogeneity, models trained in one region may perform poorly in another, and the same holds for different application tasks~\cite{jiang2019survey}. Large language models have demonstrated that with sufficient training data, it's possible to train a general model capable of performing well across diverse geospatial regions and various downstream geoscience tasks. For example, in remote sensing, where different applications may share common spectral signatures in satellite imagery, it becomes feasible to train a single neural network model to extract shared feature representations from the vast Earth imagery big data.

\subsection{Illustrative Transformative Applications}

\emph{Foundation model for Earth imagery big data}: A foundation model is a large deep neural network model trained on a vast quantity of data at scale (often by self-supervised or semi-supervised learning) such that it can be easily adapted to a wide range of downstream tasks~\cite{bommasani2021opportunities}. For example, a visual foundation model such as DINOv2 can produce all-purpose features from an image that work across image distributions and tasks without the need for  finetuning~\cite{oquab2023dinov2}. Given the petabytes of Earth imagery collected by NASA and other commercial satellite companies, there is growing interest in developing such foundation models in Earth sciences~\cite{nasafoundation}. Such an Earth imagery foundation model is transformative to geoscience research. For many years, a geoscience researcher has to first handcraft spectral signatures for specific tasks. Even with the popularity of deep convolutional neural networks, people still need to collect a large number of training labels for different application~\cite{karpatne2016monitoring}. With a foundation model, a user can potentially do any downstream task in any region with only a small number of labels for fine-tuning. However, several caveats exist. Thematic classes in geoscience are often not limited to discrete objects (e.g., buildings, roads, cars). Techniques for learning general features for object and image classes in computer vision may not be effective for land cover and land use classes. Moreover, scientists often prefer a model they can understand, interpret, and trust.

\emph{Spatiotemporal AI surrogate for Earth system models:} Another potential application is the development of surrogate models for numerical simulation models in Earth sciences (e.g., weather forecasting, hydrology, oceanography)~\cite{kurth2023fourcastnet,shi2022gnn}. For example, coastal circulation refers to the movement and circulation patterns of water in the coastal regions of oceans and seas~\cite{kumar2012implementation, olabarrieta2011wave}. Effective and efficient coastal circulation modeling plays a crucial role in building early warning and forecasting systems for coastal hazards (e.g., floods and storm surges, marine heatwaves, red tides)~\cite{olabarrieta2012ocean, hsu2023total}. Such models often contain hundreds of thousands of lines of code that run hundreds of CPUs in a high-performance computing platform. For instance, the Regional Ocean Model System (ROMS) is a free-surface, terrain-following, primitive equations ocean model widely used by the scientific community to create high-resolution simulations of ocean and coastal processes, including circulation, temperature, salinity, and other variables~\cite{roms}.
Very often, scientists need to run the simulation models multiple times for parameter calibration and uncertainty quantification. With the popularity of deep learning and GPU computational architectures, scientists have become increasingly interested in training machine learning surrogates for numerical models based on a vast number of simulation data~\cite{he2023hierarchical}. The advantage is that once a model is well-trained, it can run on a desktop or even a laptop with one or a few GPUs. This is very important for researchers and stakeholders in developing countries without access to significant computational resources.


\section{Unique Challenges}
Although STBD shares similar formats with image and video data (spatial raster resembles imagery, and temporal snapshots resemble video frames), it poses some unique challenges for deep learning technologies, as summarized below. 

First, as the space is continuous, the input size of a spatial raster map can be easily many thousand by many thousand pixels. This is much larger than the common input shapes of existing deep learning models, which are a few hundred by a few hundred pixels. Moreover, there exist complex long-range spatial dependencies across multiple spatial scales. Thus, modeling the complex spatial dependency within a large input map poses computational challenges such as a large GPU memory footprint. 

Second, spatial data contain multiple map layers co-registered in the same spatial reference system. Each individual map layer can have a distinct spatial resolution and format (vector versus raster) and diverse noise and errors. For example, in water forecasting, the input feature layers can include satellite images for land cover maps and climate model projections, sensor observations at weather stations, river network polylines, and lake polygons.

Third, training deep neural networks requires a large amount of ground truth labels (e.g., ImageNet). Unfortunately, in spatiotemporal applications, collecting ground truth is often slow, tedious, and expensive due to the need for field surveys or visually interpreting high-resolution Earth imagery. 

Fourth, spatiotemporal problems are often highly interdisciplinary and require the integration of deep learning with physical knowledge~\cite{karpatne2022knowledge}. This is because machine learning problems are often under-constrained. Without incorporating domain physics, pure data-driven deep learning can easily learn spurious relationships that look deceptively good on the training data (even on validation and test data) but do not generalize well to complex real-world scenarios. Indeed, how to integrate universal physics into machine learning to enhance model generalizability for a new region with limited labels has been listed as an open research question by the CRA's report on \emph{20-Year Roadmap for Artificial Intelligence}~\cite{gil201920}.

Finally, compared with many physics-based models in scientific domains (e.g., hydrology, oceanography), deep neural network models are black-box in nature with limited explainability, which is a concern as the end goal is to advance knowledge and understanding of a subject. Moreover, as it is difficult to explain why a deep neural network works, it is also hard to determine when it will fail. The issue becomes more serious when it comes to model generalizability in a complex real-world environment. Thus, uncertainty quantification is of great importance~\cite{he2023survey}.

\section{Future Research Directions}
Addressing the challenges listed above requires additional research in the design of novel deep-learning models and learning algorithms for STBD. Several future research directions are listed here. 

\subsection{Learning complex spatial dependency in continuous space}
In traditional spatiotemporal data mining, the common practice of prediction is to first establish a spatial neighborhood graph and then build a model to capture the dependency structure within the neighborhood graph~\cite{jiang2012learning}. For deep learning models, neighborhood dependency can be learned by convolutional filters. However, for long-range dependency beyond the neighborhood, convolutional operations are no longer sufficient. One potential direction is to leverage the more recent attention mechanisms that are popular in vision transformers~\cite{han2022survey}.
However, due to the high computational costs, we cannot compute all pair attention across all pixels (or even all pixel blocks). This problem is even worse for a large spatial raster with many thousands by many thousands of pixels. One potential direction is to capture such long-range spatial dependency at different scales with some model components to learn spatial dependency within smaller patches and other model components to learn the dependency across patches. Another non-trivial issue exists for a large number of spatial point samples in continuous space. This requires new positional encoding and efficient design of self-attention layers across a large number of point samples.

\subsection{Weakly supervised spatiotemporal deep learning}
Although there is a paucity of high-quality ground truth in spatiotemporal applications, there are often abundant weak labels that are sparse, coarse, and noisy. For example, in flood mapping applications, the ground truth from water sensors can be spatially sparse, and human-annotated flood extent boundary polygons on high-resolution imagery are also imprecise (e.g., due to obscurity from tree canopies). Similarly, in land cover mapping at a large scale (e.g., continental or global), the weak labels can come from existing classification products based on low-resolution satellite imagery. 
Addressing this issue requires novel weak supervised learning~\cite{zhou2018brief} algorithms that can train effective spatiotemporal deep learning models from sparse, coarse, and noisy labels. For sparse labels, semi-supervised learning or domain-knowledge-guided learning can potentially help fill the gap of incomplete labels. For coarse training labels, the learning algorithms may require inferring the hidden true label in a higher resolution iteratively while training an end model. The process may involve multi-instance learning. For noisy labels, a further noise model needs to be assumed to capture the probabilistic relationship between observed noisy labels and the hidden true labels to facilitate label inference during model training. The process becomes more complex if the input labels are in the vector format (e.g., points, polylines, and polygons) while the input features are in spatial raster imagery~\cite{he2022quantifying,jiang2022weakly}.

\subsection{Knowledge-guided spatiotemporal deep learning} 
There are different sources of domain physical knowledge. 
One source is symbolic knowledge in a geographical knowledge graph or knowledge base (with spatial logic rules). To integrate such knowledge into spatiotemporal deep learning requires neurosymbolic models. One potential challenge is that the underlying phenomenon is in continuous space without explicit object definitions. How to ground a geospatial knowledge base (logic rules) into discrete locations involves a trade-off between computational efficiency and spatial granularity~\cite{xu2023spatial}. 
Another source of geographical knowledge is numerical physical models based on partial differential equations (PDEs). Examples include weather forecasting models, ocean circulation models, and hydrologic models. There are already extensive studies in using deep neural networks to solve PDEs, such as physics-informed neural networks (PINNs)~\cite{raissi2019physics} and neural operator learning~\cite{kovachki2023neural}.  PINN aims to train a neural network as a function of spatial and temporal coordinates. Samples on the initial and boundary conditions are considered as ground truth and other randomly sampled locations in between are considered as regularization terms based on the PDE constraint. Neural operator learning aims to train a neural network as an operator for a family of PDEs, taking in samples on the initial and boundary conditions as network inputs and producing a solution function as the neural network outputs. The advantage of PINNs is that it does not require solving a PDE first, as the neural network training process is solving the PDE, possibly at a faster speed due to the efficient GPU parallelization and automatic differentiation.  However, the neural network itself is still a simple function of coordinates, often multilayer perceptrons, without explicitly encoding the spatial dependency between different sample locations.
Training a PINN for complex real-world PDEs can be challenging. The network needs to be retrained for a new PDE instance. As for neural operator learning, it only requires model training once for all but requires a significant amount of simulations from numerical solvers. It is often challenging to collect a large number of training instances for real-world scenarios due to the high time costs of solving those PDEs. 

\subsection{Model generalizability across heterogeneity}
There are several issues to be addressed here. First, to address the challenge of spatial and temporal heterogeneity, one potential idea is to design new learning algorithms to be able to train an ensemble of models (each base model is trained for one homogeneous sub-area). For example, meta-transfer learning could be used to effectively train a shared module across different domain tasks~\cite{sun2019meta}. For spatial problems, the definition of different domains (e.g., homogeneous sub-regions) can be implicit. Several works have been done to address this issue~\cite{jiang2017spatial,jiang2019spatial}. Another more challenging issue is the out-of-distribution (OOD) scenario. A well-trained model with high training and validation accuracy can still show poor performance when being deployed in a complex real-world environment. To address this issue, the model needs to be able to identify when a test case is an OOD case and accurately quantify the uncertainty of its prediction, e.g., based on sample distance or density in feature space or geographical space~\cite{he2022quantifying}.

\section*{Acknowledgment}
This material is based upon work supported by the National Science Foundation (NSF) under Grant No. IIS-2147908, IIS-2207072, CNS-1951974, OAC-2152085.

\bibliographystyle{IEEEtran}
\bibliography{ref,maitane,jiang_pub}

\begin{thebibliography}{10}
\providecommand{\url}[1]{#1}
\csname url@samestyle\endcsname
\providecommand{\newblock}{\relax}
\providecommand{\bibinfo}[2]{#2}
\providecommand{\BIBentrySTDinterwordspacing}{\spaceskip=0pt\relax}
\providecommand{\BIBentryALTinterwordstretchfactor}{4}
\providecommand{\BIBentryALTinterwordspacing}{\spaceskip=\fontdimen2\font plus
\BIBentryALTinterwordstretchfactor\fontdimen3\font minus
  \fontdimen4\font\relax}
\providecommand{\BIBforeignlanguage}[2]{{%
\expandafter\ifx\csname l@#1\endcsname\relax
\typeout{** WARNING: IEEEtran.bst: No hyphenation pattern has been}%
\typeout{** loaded for the language `#1'. Using the pattern for}%
\typeout{** the default language instead.}%
\else
\language=\csname l@#1\endcsname
\fi
#2}}
\providecommand{\BIBdecl}{\relax}
\BIBdecl

\bibitem{eftelioglu2017nexus}
E.~Eftelioglu, Z.~Jiang, X.~Tang, and S.~Shekhar, ``The nexus of food, energy,
  and water resources: Visions and challenges in spatial computing,'' in
  \emph{Advances in Geocomputation: Geocomputation 2015--The 13th International
  Conference}.\hskip 1em plus 0.5em minus 0.4em\relax Springer, 2017, pp.
  5--20.

\bibitem{nasadata}
{NASA}, ``{NASA's Earth Science Data Systems (ESDS) Program},''
  \url{https://www.earthdata.nasa.gov/technology/open-science}, Last Accessed
  on August 20, 2023.

\bibitem{jiang2017spatialbig}
Z.~Jiang and S.~Shekhar, \emph{Spatial Big Data Science - Classification
  Techniques for Earth Observation Imagery}.\hskip 1em plus 0.5em minus
  0.4em\relax Springer, 2017.

\bibitem{shekhar2015spatiotemporal}
S.~Shekhar, Z.~Jiang, R.~Y. Ali, E.~Eftelioglu, X.~Tang, V.~Gunturi, and
  X.~Zhou, ``Spatiotemporal data mining: A computational perspective,''
  \emph{ISPRS International Journal of Geo-Information}, vol.~4, no.~4, pp.
  2306--2338, 2015.

\bibitem{atluri2018spatio}
G.~Atluri, A.~Karpatne, and V.~Kumar, ``Spatio-temporal data mining: A survey
  of problems and methods,'' \emph{ACM Computing Surveys (CSUR)}, vol.~51,
  no.~4, pp. 1--41, 2018.

\bibitem{jiang2019survey}
Z.~Jiang, ``A survey on spatial prediction methods,'' \emph{IEEE Transactions
  on Knowledge \& Data Engineering}, vol.~31, no.~09, pp. 1645--1664, 2019.

\bibitem{sharma202221}
A.~Sharma, Z.~Jiang, and S.~Shekhar, ``Spatiotemporal data mining,''
  \emph{Handbook of Spatial Analysis in the Social Sciences}, p. 352, 2022.

\bibitem{cressie2015statistics}
N.~Cressie and C.~K. Wikle, \emph{Statistics for spatio-temporal data}.\hskip
  1em plus 0.5em minus 0.4em\relax John Wiley \& Sons, 2015.

\bibitem{aji2013hadoop}
A.~Aji, F.~Wang, H.~Vo, R.~Lee, Q.~Liu, X.~Zhang, and J.~Saltz, ``Hadoop-gis: A
  high performance spatial data warehousing system over mapreduce,'' in
  \emph{Proceedings of the VLDB endowment international conference on very
  large data bases}, vol.~6, no.~11.\hskip 1em plus 0.5em minus 0.4em\relax NIH
  Public Access, 2013.

\bibitem{eldawy2015spatialhadoop}
A.~Eldawy and M.~F. Mokbel, ``Spatialhadoop: A mapreduce framework for spatial
  data,'' in \emph{2015 IEEE 31st international conference on Data
  Engineering}.\hskip 1em plus 0.5em minus 0.4em\relax IEEE, 2015, pp.
  1352--1363.

\bibitem{yu2015geospark}
J.~Yu, J.~Wu, and M.~Sarwat, ``Geospark: A cluster computing framework for
  processing large-scale spatial data,'' in \emph{Proceedings of the 23rd
  SIGSPATIAL international conference on advances in geographic information
  systems}, 2015, pp. 1--4.

\bibitem{lecun2015deep}
Y.~LeCun, Y.~Bengio, and G.~Hinton, ``Deep learning,'' \emph{nature}, vol. 521,
  no. 7553, pp. 436--444, 2015.

\bibitem{reichstein2019deep}
M.~Reichstein, G.~Camps-Valls, B.~Stevens, M.~Jung, J.~Denzler, N.~Carvalhais,
  and f.~Prabhat, ``Deep learning and process understanding for data-driven
  earth system science,'' \emph{Nature}, vol. 566, no. 7743, pp. 195--204,
  2019.

\bibitem{jumper2021highly}
J.~Jumper, R.~Evans, A.~Pritzel, T.~Green, M.~Figurnov, O.~Ronneberger,
  K.~Tunyasuvunakool, R.~Bates, A.~{\v{Z}}{\'\i}dek, A.~Potapenko
  \emph{et~al.}, ``Highly accurate protein structure prediction with
  alphafold,'' \emph{Nature}, vol. 596, no. 7873, pp. 583--589, 2021.

\bibitem{yin2021deep}
X.~Yin, G.~Wu, J.~Wei, Y.~Shen, H.~Qi, and B.~Yin, ``Deep learning on traffic
  prediction: Methods, analysis, and future directions,'' \emph{IEEE
  Transactions on Intelligent Transportation Systems}, vol.~23, no.~6, pp.
  4927--4943, 2021.

\bibitem{brown2020language}
T.~Brown, B.~Mann, N.~Ryder, M.~Subbiah, J.~D. Kaplan, P.~Dhariwal,
  A.~Neelakantan, P.~Shyam, G.~Sastry, A.~Askell \emph{et~al.}, ``Language
  models are few-shot learners,'' \emph{Advances in neural information
  processing systems}, vol.~33, pp. 1877--1901, 2020.

\bibitem{bommasani2021opportunities}
R.~Bommasani, D.~A. Hudson, E.~Adeli, R.~Altman, S.~Arora, S.~von Arx, M.~S.
  Bernstein, J.~Bohg, A.~Bosselut, E.~Brunskill \emph{et~al.}, ``On the
  opportunities and risks of foundation models,'' \emph{arXiv e-prints}, pp.
  arXiv--2108, 2021.

\bibitem{oquab2023dinov2}
M.~Oquab, T.~Darcet, T.~Moutakanni, H.~Vo, M.~Szafraniec, V.~Khalidov,
  P.~Fernandez, D.~Haziza, F.~Massa, A.~El-Nouby \emph{et~al.}, ``Dinov2:
  Learning robust visual features without supervision,'' \emph{arXiv preprint
  arXiv:2304.07193}, 2023.

\bibitem{nasafoundation}
{NASA}, ``{NASA and IBM Openly Release Geospatial AI Foundation Model for NASA
  Earth Observation Data },''
  \url{https://www.earthdata.nasa.gov/news/impact-ibm-hls-foundation-model},
  Last Accessed on August 20, 2023.

\bibitem{karpatne2016monitoring}
A.~Karpatne, Z.~Jiang, R.~R. Vatsavai, S.~Shekhar, and V.~Kumar, ``Monitoring
  land-cover changes: A machine-learning perspective,'' \emph{IEEE Geoscience
  and Remote Sensing Magazine}, vol.~4, no.~2, pp. 8--21, 2016.

\bibitem{kurth2023fourcastnet}
T.~Kurth, S.~Subramanian, P.~Harrington, J.~Pathak, M.~Mardani, D.~Hall,
  A.~Miele, K.~Kashinath, and A.~Anandkumar, ``Fourcastnet: Accelerating global
  high-resolution weather forecasting using adaptive fourier neural
  operators,'' in \emph{Proceedings of the Platform for Advanced Scientific
  Computing Conference}, 2023, pp. 1--11.

\bibitem{shi2022gnn}
N.~Shi, J.~Xu, S.~W. Wurster, H.~Guo, J.~Woodring, L.~P. Van~Roekel, and H.-W.
  Shen, ``Gnn-surrogate: A hierarchical and adaptive graph neural network for
  parameter space exploration of unstructured-mesh ocean simulations,''
  \emph{IEEE Transactions on Visualization and Computer Graphics}, vol.~28,
  no.~6, pp. 2301--2313, 2022.

\bibitem{kumar2012implementation}
N.~Kumar, G.~Voulgaris, J.~C. Warner, and M.~Olabarrieta, ``Implementation of
  the vortex force formalism in the coupled ocean-atmosphere-wave-sediment
  transport (coawst) modeling system for inner shelf and surf zone
  applications,'' \emph{Ocean Modelling}, vol.~47, pp. 65--95, 2012.

\bibitem{olabarrieta2011wave}
M.~Olabarrieta, J.~C. Warner, and N.~Kumar, ``Wave-current interaction in
  willapa bay,'' \emph{Journal of Geophysical Research: Oceans}, vol. 116, no.
  C12, 2011.

\bibitem{olabarrieta2012ocean}
M.~Olabarrieta, J.~C. Warner, B.~Armstrong, J.~B. Zambon, and R.~He,
  ``Ocean--atmosphere dynamics during hurricane ida and nor’ida: An
  application of the coupled ocean--atmosphere--wave--sediment transport
  (coawst) modeling system,'' \emph{Ocean Modelling}, vol.~43, pp. 112--137,
  2012.

\bibitem{hsu2023total}
C.-E. Hsu, K.~Serafin, X.~Yu, C.~Hegermiller, J.~C. Warner, and M.~Olabarrieta,
  ``Total water levels along the south atlantic bight during three along-shelf
  propagating tropical cyclones: relative contributions of storm surge and wave
  runup,'' \emph{Natural Hazards and Earth System Sciences Discussions}, vol.
  2023, pp. 1--31, 2023.

\bibitem{roms}
D.~O.~M. Group, ``{Regional Ocean Modeling System (ROMS)},''
  \url{https://www.myroms.org/}, 2023.

\bibitem{he2023hierarchical}
W.~He, Z.~Jiang, T.~Xiao, Z.~Xu, S.~Chen, R.~Fick, M.~Medina, and C.~Angelini,
  ``A hierarchical spatial transformer for massive point samples in continuous
  space,'' in \emph{Proceedings of the 2023 Conference on Neural Information
  Processing Systems (NeurIPS)}, 2023.

\bibitem{karpatne2022knowledge}
A.~Karpatne, R.~Kannan, and V.~Kumar, \emph{Knowledge Guided Machine Learning:
  Accelerating Discovery Using Scientific Knowledge and Data}.\hskip 1em plus
  0.5em minus 0.4em\relax CRC Press, 2022.

\bibitem{gil201920}
Y.~Gil and B.~Selman, ``A 20-year community roadmap for artificial intelligence
  research in the us,'' \emph{arXiv preprint arXiv:1908.02624}, 2019.

\bibitem{he2023survey}
W.~He and Z.~Jiang, ``A survey on uncertainty quantification methods for deep
  neural networks: An uncertainty source perspective,'' \emph{arXiv preprint
  arXiv:2302.13425}, 2023.

\bibitem{jiang2012learning}
Z.~Jiang, S.~Shekhar, P.~Mohan, J.~Knight, and J.~Corcoran, ``Learning spatial
  decision tree for geographical classification: a summary of results,'' in
  \emph{Proceedings of the 20th International Conference on Advances in
  Geographic Information Systems}, 2012, pp. 390--393.

\bibitem{han2022survey}
K.~Han, Y.~Wang, H.~Chen, X.~Chen, J.~Guo, Z.~Liu, Y.~Tang, A.~Xiao, C.~Xu,
  Y.~Xu \emph{et~al.}, ``A survey on vision transformer,'' \emph{IEEE
  transactions on pattern analysis and machine intelligence}, vol.~45, no.~1,
  pp. 87--110, 2022.

\bibitem{zhou2018brief}
Z.-H. Zhou, ``A brief introduction to weakly supervised learning,''
  \emph{National science review}, vol.~5, no.~1, pp. 44--53, 2018.

\bibitem{he2022quantifying}
W.~He, Z.~Jiang, M.~Kriby, Y.~Xie, X.~Jia, D.~Yan, and Y.~Zhou, ``Quantifying
  and reducing registration uncertainty of spatial vector labels on earth
  imagery,'' in \emph{Proceedings of the 28th ACM SIGKDD Conference on
  Knowledge Discovery and Data Mining}, 2022, pp. 554--564.

\bibitem{jiang2022weakly}
Z.~Jiang, W.~He, M.~S. Kirby, A.~M. Sainju, S.~Wang, L.~V. Stanislawski, E.~J.
  Shavers, and E.~L. Usery, ``Weakly supervised spatial deep learning for earth
  image segmentation based on imperfect polyline labels,'' \emph{ACM
  Transactions on Intelligent Systems and Technology (TIST)}, vol.~13, no.~2,
  pp. 1--20, 2022.

\bibitem{xu2023spatial}
Z.~Xu, T.~Xiao, W.~He, Y.~Wang, and Z.~Jiang, ``Spatial knowledge-infused
  hierarchical learning: An application in flood mapping on earth imagery,'' in
  \emph{The 31st ACM SIGSPATIAL International Conference on Advances in
  Geographic Information Systems (GIS)}, 2023.

\bibitem{raissi2019physics}
M.~Raissi, P.~Perdikaris, and G.~E. Karniadakis, ``Physics-informed neural
  networks: A deep learning framework for solving forward and inverse problems
  involving nonlinear partial differential equations,'' \emph{Journal of
  Computational physics}, vol. 378, pp. 686--707, 2019.

\bibitem{kovachki2023neural}
N.~B. Kovachki, Z.~Li, B.~Liu, K.~Azizzadenesheli, K.~Bhattacharya, A.~M.
  Stuart, and A.~Anandkumar, ``Neural operator: Learning maps between function
  spaces with applications to pdes.'' \emph{J. Mach. Learn. Res.}, vol.~24,
  no.~89, pp. 1--97, 2023.

\bibitem{sun2019meta}
Q.~Sun, Y.~Liu, T.-S. Chua, and B.~Schiele, ``Meta-transfer learning for
  few-shot learning,'' in \emph{Proceedings of the IEEE/CVF conference on
  computer vision and pattern recognition}, 2019, pp. 403--412.

\bibitem{jiang2017spatial}
Z.~Jiang, Y.~Li, S.~Shekhar, L.~Rampi, and J.~Knight, ``Spatial ensemble
  learning for heterogeneous geographic data with class ambiguity: A summary of
  results,'' in \emph{Proceedings of the 25th ACM SIGSPATIAL international
  conference on advances in geographic information systems}, 2017, pp. 1--10.

\bibitem{jiang2019spatial}
Z.~Jiang, A.~M. Sainju, Y.~Li, S.~Shekhar, and J.~Knight, ``Spatial ensemble
  learning for heterogeneous geographic data with class ambiguity,'' \emph{ACM
  Transactions on Intelligent Systems and Technology (TIST)}, vol.~10, no.~4,
  pp. 1--25, 2019.

\end{thebibliography}
\end{document}